\documentclass[conference]{IEEEtran}
\IEEEoverridecommandlockouts
\usepackage{cite}
\usepackage{amsmath,amssymb,amsfonts}
\usepackage{amsmath,epsfig,algorithmic,algorithm}
\usepackage{amssymb}
\usepackage{algorithmic}
\usepackage{graphicx}
\usepackage{textcomp}
\usepackage{xcolor}
\usepackage{multicol}
\usepackage{graphicx}
\usepackage{comment}
\newcommand{\cdj}[1]{\textcolor{black}{#1}}
\newcommand{\scc}[1]{\textcolor{black}{#1}}

\DeclareMathOperator*{\argmin}{arg\,min}

\newcommand\bfootnote[1]{%
  \begingroup
  \renewcommand\thefootnote{}\footnote{#1}%
  \addtocounter{footnote}{-1}%
  \endgroup
}

\def\BibTeX{{\rm B\kern-.05em{\sc i\kern-.025em b}\kern-.08em
    T\kern-.1667em\lower.7ex\hbox{E}\kern-.125emX}}
\begin{document}

\title{Differentially Private Generative Adversarial Networks with Model Inversion
}

\author{\IEEEauthorblockN{Dongjie Chen}
\IEEEauthorblockA{
\textit{University of California}\\
Davis, CA \\
cdjchen@ucdavis.edu}\\
\and
\IEEEauthorblockN{Sen-ching Samson Cheung}
\IEEEauthorblockA{
\textit{University of Kentucky}\\
Lexington, KY \\
sccheung@ieee.org}\\
\and
\IEEEauthorblockN{Chen-Nee Chuah}
\IEEEauthorblockA{
\textit{University of California}\\
Davis, CA\\
chuah@ucdavis.edu}\\
\and
\IEEEauthorblockN{Sally Ozonoff}
\IEEEauthorblockA{
\textit{University of California}\\
Davis, CA\\
sozonoff@ucdavis.edu}\\
}

\maketitle

\begin{abstract}
To protect sensitive data in training a Generative Adversarial Network (GAN), the standard approach is to use differentially private (DP) stochastic gradient descent method in which controlled noise is added to the gradients. The quality of the output synthetic samples can be adversely affected and the training of the network may not even converge in the presence of these noises. We propose Differentially Private Model Inversion (DPMI) method where the private data is first mapped to the latent space via a public generator, followed by a lower-dimensional DP-GAN with better convergent properties. Experimental results on standard datasets CIFAR10 and SVHN as well as on a facial landmark dataset for Autism screening show that our approach outperforms the standard DP-GAN method based on Inception Score, Frechet Inception Distance, and classification accuracy under the same privacy guarantee. 
\end{abstract}


\bfootnote{Best Student Paper Award of 13th IEEE International Workshop on Information Forensics and Security (WIFS 2021), Montpellier, France. © 2021 IEEE.

DOI: 10.1109/WIFS53200.2021.9648378 }

\section{Introduction}
With the advances in digital information collection techniques, governments and private enterprises can easily collect sensitive information about anyone in the society. As such, database privacy has become an important topic for legal, ethical, and technical reasons. According to \cite{dwork2014algorithmic}, database privacy can be achieved via one of two paradigms. In an interactive setting, a data curator collects data from individuals and provides privacy preserving interfaces for analysts to query data. In a non-interactive setting, a curator releases a ``sanitized'' or ``synthetic'' version of data to protect privacy and ensure utility. After the data is released, the curators will not modify the data again and may delete the original data. In the age of deep learning, a common way for non-interactive data release is to generate synthetic data using a Generative Adversarial Networks (GANs) trained on the original private data. 

GANs have gained significant popularity since its invention in \cite{goodfellow2014generative} and have been used in many applications including 
image-to-image translation~\cite{isola2017image}, face image synthesis~\cite{zhang2019self}
, and image inpainting~\cite{pathak2016context}. As GANs gain popularity, there are also increasing privacy concerns on whether GANs can compromise the privacy of the original training data~\cite{fan2020survey}, particularly if they are used for private data release such as medical data sharing~\cite{esteban2017real}. Attack vectors on GANs such as membership inference attacks~\cite{shokri2017membership} and full reconstruction~\cite{fredrikson2015model} have been investigated. As such, it is imperative to develop robust privacy protection schemes for GANs. 

The de facto approach to provide privacy guarantee in machine learning is through the differential privacy (DP) framework~\cite{dwork2014algorithmic}. Two different DP schemes have been proposed for GANs: Private Aggregation of Teacher Ensembles (PATE) and Differentially Private Stochastic Gradient Descent (DPSGD). PATE-GAN, proposed in ~\cite{jordon2018pate}, achieves differential privacy by training multiple teacher discriminators using disjoint parts of the training set and one student discriminator. The requirement of using many teacher discriminators trained on different parts of the training set makes PATE-GAN challenging to scale to high-dimensional datasets with limited number of training samples~\cite{chen2020gs}.

DPSGD, originally proposed in~\cite{abadi2016deep}, aims at making the SGD process differentially private by clipping and adding random noise to the aggregated gradients. The usage of DPSGD goes beyond GAN and has been applied to many SGD-based deep-learning  techniques~\cite{xie2018differentially,xu2019ganobfuscator,torkzadehmahani2019dp,augenstein2019generative}. The unique challenge of DPSGD-based GANs is its difficulty in training - due to the game-theoretic formulation of the cost function, the training process of a complicated GAN does not necessarily converge or converges to a noisy equilibrium, resulting in mode collapse and/or poor synthetic sample quality~\cite{fan2020survey}. A common remedy is to repeat the training process until convergence or depletion of the privacy budget~\cite{beaulieu2019privacy}, \scc{but such a method does not necessarily help the convergence of the training process. As such, DPSGD has not been widely tested on high dimensional multimedia datasets beyond simple test cases such as MNIST.}

\begin{figure}[h!]
\begin{minipage}[b]{0.45\linewidth}
  \centering
  \includegraphics[width=\linewidth]{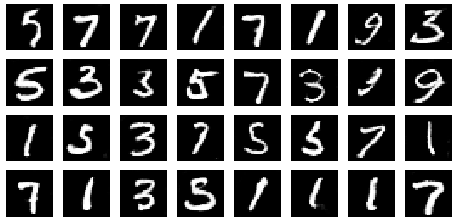}
  \centerline{(a)}\medskip
\end{minipage}
\hfill
\begin{minipage}[b]{0.45\linewidth}
  \centering
  \includegraphics[width=\linewidth]{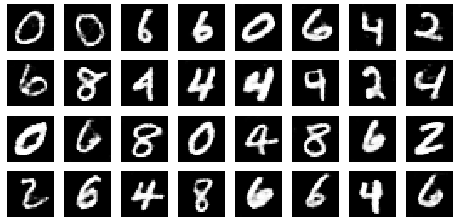}
  \centerline{(b)}\medskip
\end{minipage}

\begin{minipage}[b]{0.45\linewidth}
  \centering
  \includegraphics[width=\linewidth]{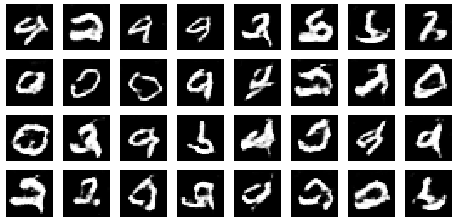}
  \centerline{(c)}\medskip
\end{minipage}
\hfill
\begin{minipage}[b]{0.45\linewidth}
  \centering
  \includegraphics[width=\linewidth]{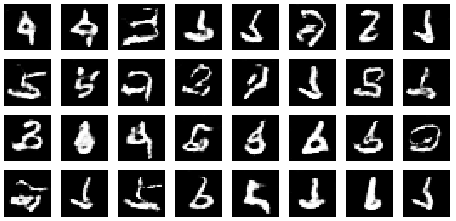}
  \centerline{(d)}\medskip
\end{minipage}
\caption{(a) Images generated by a generator $G$ trained odd digits only;  (b) images generated by $G$ through model inversion with even digits; (c) even-digit images generated by the proposed DPMI method without DP ($\epsilon=\infty$); (d) even-digit images generated by DPMI with $\epsilon=10$.}
\label{fig:mnistmi}
\end{figure}

\begin{figure}[h!]
\centering
\centering
\includegraphics[width=0.7\linewidth]{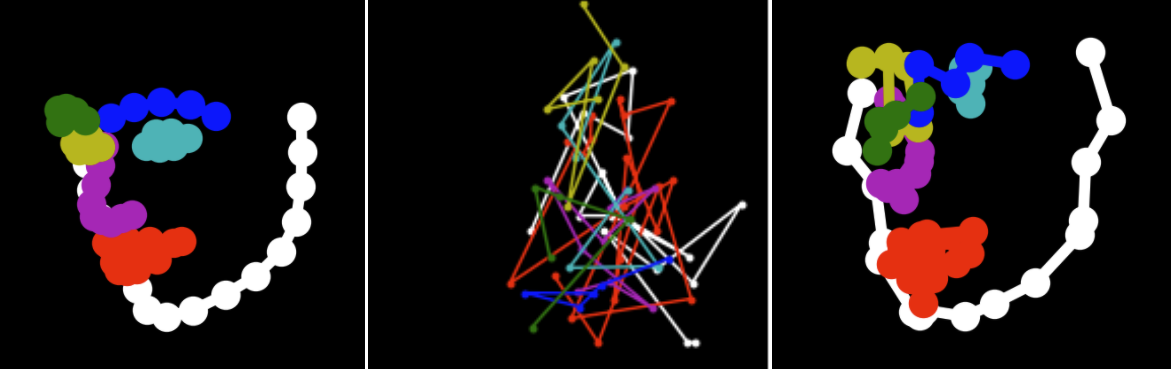}
\caption{\cdj{From the left to the right: facial landmarks from the real ASDF dataset, synthetic facial landmarks generated by DP-GAN with $\epsilon=10$, synthetic facial landmarks generated by proposed DPMI with $\epsilon=10$.} }
\label{fig:openface_intro}
\end{figure}

In this paper, we propose a novel approach called Differentially Private Model Inversion (DPMI) to improve DPSGD-based GANs. Due to the difficulty in training complex DP-GANs, DPMI first trains the same high-dimensional GAN (latent dimensional space $\mathbb{R}^d$) but using only public data, thereby negating the need of using any DP noise. To align with the statistics of the private data, we use the trained generator $G_{p}$ and identify a group of private latent vectors $z_{s}\in \mathbb{R}^d$ such that the similarity between $G_{p}(z_{s})$ and the private data is maximized. To protect the privacy of $z_{s}$, we train another DP-GAN with generator $G_{ds}$ that maps lower-dimensional latent vectors $z_{ds}\in\mathbb{R}^l$ with $l<d$ to $z_{s}$. Due to a lower dimension and less complex network structure, this smaller DP-GAN can be trained with better stability than the original DP-GAN at the same privacy level~\cite{neyshabur2017stabilizing}. Our experimental results show that synthesizing data using $G_{p}(G_{ds}(z_{ds}))$ with random $z_{ds}\in\mathbb{R}^l$ is differentially private, more stable to train, and can produce higher subsequent learning performance than training a single DP-GAN. 

In Figure \ref{fig:mnistmi}, we show how DPMI works using the odd digits from MNIST as public data and the even digits as private data. Figure \ref{fig:mnistmi}(a) shows that $G_{p}$ can only generate odd digits. After model inversion, Figure \ref{fig:mnistmi}(b) shows high-quality even digits $G_{p}(z_{s})$. Figure \ref{fig:mnistmi}(c) and (d) show the even digits generated $G_{p}(G_{ds}(z_{ds}))$ with no DP noise and with DP noise at $\epsilon=10$ respectively. Though not of the highest quality, all even digits were sufficiently hallucinated using a GAN trained only on odd digits. 

\cdj{To further illustrate how DPMI performs on a privacy-sensitive real-world application, we have conducted experiments on a medical dataset. The Autism Spectrum Disorder Face (ASDF) dataset contains facial landmarks of children extracted from videos used for ASD screening. At $\epsilon=10$, Figure \ref{fig:openface_intro} shows that DPMI can generate realistic facial landmarks while the traditional DP-GAN is not able to converge.}

The remainder of this paper is organized as follows. In Section \ref{sec:background}, we review the fundamentals of GAN and differential privacy. Section \ref{sec:method} describes our proposed method and Section \ref{sec:experiments} compares the performance of the proposed methods with other DP-GANs under various conditions. We conclude the paper in Section \ref{sec:conclusion}.

\section{Background}
\label{sec:background}

In this section, we review some basic concepts from GAN and differential privacy. 

\subsection{GAN and WGAN}
A typical GAN consists of two networks: a generator $G(z) \in X$ that maps a latent vector $z \in \mathbb{R}^d$ to the target image space $X$, and a discriminator $C(x) \in \{0,1\}$ that determines if an image input $x \in X$ looks real (1) or fake (0). The distribution $P_Z$ of the latent vector is usually set to be a separable $d$-dimensional Gaussian distribution. Assume that the real data comes from a distribution $P_X$, the goal of training a GAN is to find $G$ and $C$ in a two-player minmax game solving the following optimization problem:
\begin{eqnarray}
\min_{G} \max_{C} \mathbb{E}_{x \sim P_X}[\log(C(x))] 
+\mathbb{E}_{z \sim P_Z}[\log(1-C(G(z)))] 
\end{eqnarray}
Wasserstein GAN or WGAN improves the regulation of the original GAN by using the Wasserstein distance between the distributions of the latent vectors and the real images ~\cite{arjovsky2017wasserstein}. Given a parametrized family of $K$-lipschitz functions $\left\{f_{w}(x)\right\}_{w\in W}$, the optimization can be approximately solved by the following value function:
\begin{eqnarray}
\min_{G} \max_{w \in W} \mathbb{E}_{x \sim P_X}\left[f_{w}(x)\right]-\mathbb{E}_{z \sim P_Z}\left[f_{w}(G(z))\right]
\end{eqnarray}
In this paper, we use the improved version of WGAN from \cite{gulrajani2017improved} to generate all our images. 

\subsection{Differential privacy}

A randomized mechanism $M$ applied onto a database $D$ is called differentially private if the output of any query will not change significantly when replacing $D$ with a neighboring database $D'$ that differs from $D$ by at most one data record. Specifically, a differentially private mechanism can be defined as follows: 
\newtheorem{mydef}{Definition}
\begin{mydef}
(Differential Privacy) A randomized mechanism $M$ is $(\epsilon,\delta)$-differential privacy if any output set $S$ and any neighboring databases $D$ and $D'$ satisfy the followings:
\begin{equation}
\mathrm{P}(\mathcal{M}(D) \in S) \leq e^{\epsilon} \cdot \mathrm{P}\left(\mathcal{M}\left(D^{\prime}\right) \in S\right)+\delta
\end{equation}
\end{mydef}
For $\delta=0$, a smaller $\epsilon$ results in a tighter bound with $\epsilon=0$ represents perfect privacy. A small positive $\delta$ allows small possibility of failure in exchange of more flexible mechanism designs. A nice property of DP mechanism, which we state here without proof, is that any post-processing applied to a DP mechanism is automatically DP: 
\newtheorem{mydef2}{Theorem} 
\begin{mydef2}
(Post-processing) Given an arbitrary mapping $f: R \rightarrow R^{\prime}$ and an $(\epsilon,\delta)$-differentially private mechanism $\mathcal{M}: D \rightarrow R$, $f \circ \mathcal{M}: D \rightarrow R^{\prime}$ is $(\epsilon,\delta)$-differentially private. \label{thm:post}
\end{mydef2}

To provide a DP mechanism for training of a deep network on private data, the most straightforward approach is to use DPSGD~\cite{xie2018differentially}, of which the key steps are summarized below:
\begin{enumerate}
    \item Randomly sample a mini-batch of private training data and compute the empirical loss and gradients.
    \item The magnitude of the gradient function for each training sample is clipped at a clipping bound $d$.
    \item Based on the target $\epsilon$ and $\delta$, a privacy accountant mechanism such as \cite{mironov2017renyi} determines the number of iterations and the variance of independent zero-mean Gaussian noise to be added to the aggregate gradient. 
    \item Update the weights of the network using the noisy gradient until the number of iteration hits the privacy budget. 
\end{enumerate}
Note that DPSGD is applied only during the training of the discriminator as the generator does not have any access to the private training data.

\section{Proposed DPMI Framework}
\label{sec:method}

\begin{figure}[h!]
\begin{center}
\includegraphics[width=\linewidth]{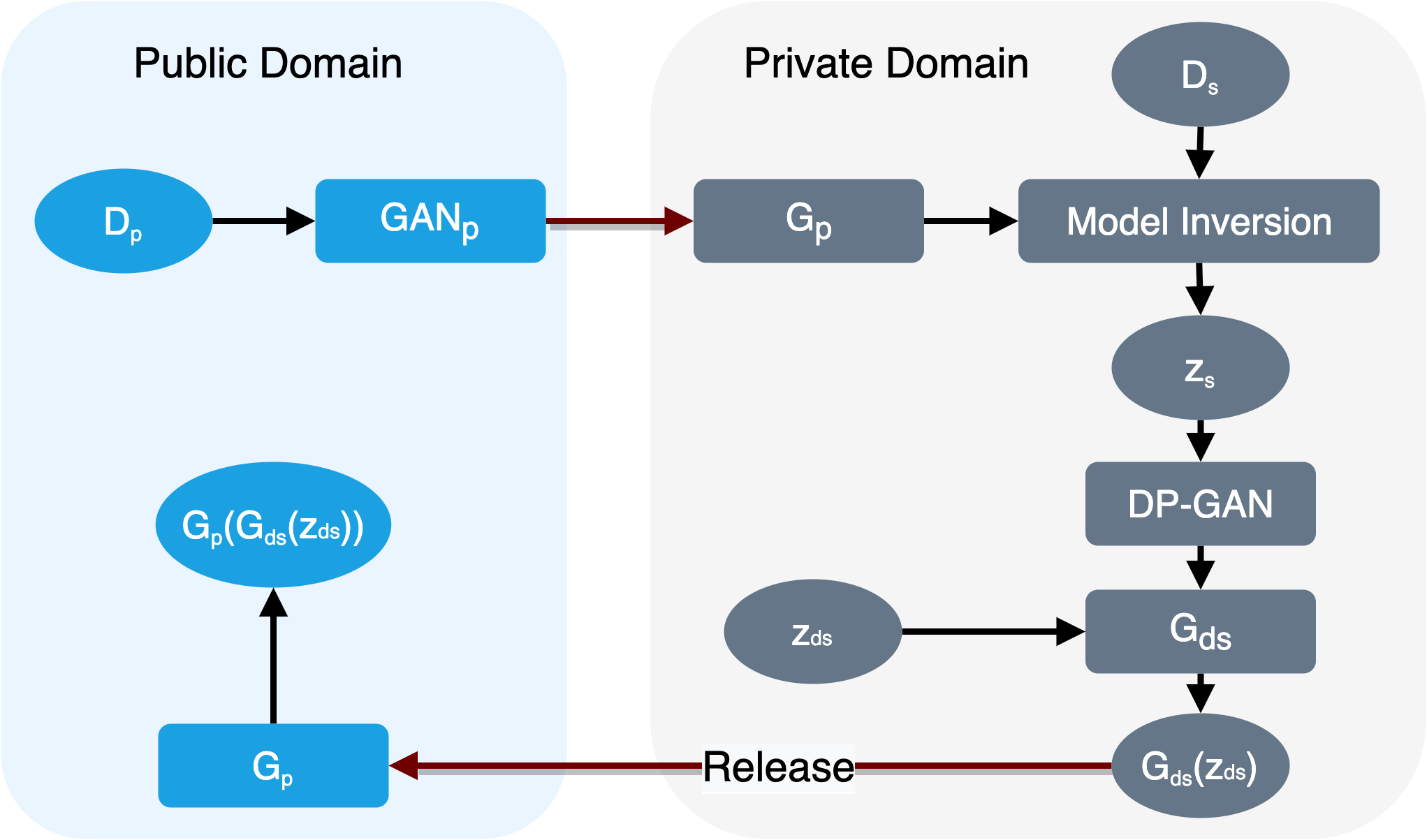}
\caption{Proposed DPMI Framework}
\label{fig:overall_frame}
\end{center}
\end{figure}

Figure~\ref{fig:overall_frame} shows the data release process of our proposed DPMI framework. In the public domain, a GAN is trained on publicly available image data $D_{p}$. The resulting trained generator $G_{p}$ is used in the private domain where a model inversion process, detailed in Section \ref{ssec:dplatentspaceGAN}, is performed to find the latent vector $z_s$ that can best approximate, via $G_{p}$, each image from the private database $D_{s}$. A lower-dimensional DP-GAN is then trained on this collection of latent vectors. The resulting DP generator $G_{ds}$ is used to publicly release a database of synthetic latent vectors that can be used to synthesize synthetic images with $G_{p}$. The privacy of this DPMI framework is discussed in Section \ref{ssec:dplatentspaceGAN}. 

\subsection{Model Inversion of GANs}
\label{ssec:modelinversion}

By training a GAN using only public data, there is no need to incorporate any DP mechanisms. This greatly mitigates the convergent problem brought on by the DP noise added to the training process. Due to the differences between the public and private images, the resulting generator will not be able to synthesize images similar to the private images. Nevertheless, with sufficient diversity in the public data, this public generator should be general enough to approximate private images given the ``right'' latent vectors. We term the process of identifying these latent vectors: $model$ $inversion$. Unlike model inversion attacks~\cite{fredrikson2015model} that are designed to learn private training data given a trained model, our model inversion process has full access to the private data and is performed entirely within the private domain.

Given a trained public generator $G_{p}$ that maps a random $d$-dimensional latent vector $z \sim P_{Z} = \mathcal{N}^d(0,I)$ to a synthetic image $G_{p}(z)$, our goal is to identify, for each private image $x_{s} \in D_{s}$, the latent vector $z_{s}$ that minimizes the mean square difference between $x_{s}$ and $G_{p}(z_{s})$. More specifically, the model inversion solves the following optimization process:
\begin{equation} 
\begin{aligned}
z_s = \argmin_{z}      \quad & ||G_{p}(z) - x_{s}||^{2}\\
\textrm{s.t.} \quad & \mbox{$P_{Z}(z) \geq P_{Z}(z_0)$ with $z_0 \sim P_{Z}$} \label{eq:MI} \\
\end{aligned}
\end{equation}
We start the model inversion process by first drawing a random sample $z_0$ from $P_{Z}$ and the constraint in (\ref{eq:MI}) is to ensure that the quality of $G_p(z_s)$ to be comparable to that of $G_p(z_0)$. This convex optimization can be easily solved via stochastic gradient ascent procedure by projecting the search trajectory back onto the convex constraint~\cite{shalev2014understanding}.

\subsection{Differentially private GAN on the latent space}
\label{ssec:dplatentspaceGAN}

With a well-trained and generalizable $G_{p}$, the model inversion can identify $z_{s} \in \mathbb{R}^d$ that can yield very good reconstruction of private images, even if there is no overlap in classes between $D_{p}$ and $D_{s}$ like the one shown in  Figure \ref{fig:mnistmi}. On the other hand, it is very difficult to make the model inversion process differentially private. The model inversion process takes a single image as the input database, which means that we must consider the significant increase in MSE when using an empty set as a neighboring database. This implies that a substantial amount of noise must be added to obfuscate between these two neighboring databases. 

In DPMI, we have adopted an alternative approach -- we treat the entire collection of $z_{s}$ as the surrogate of $D_{S}$ and train a DP-GAN to synthesize latent vectors that closely resemble the private $z_{s}$'s. Unlike the image-based DP-GAN, this latent space DP-GAN is of substantially lower dimension with a very well-defined range space as they are all originated from $\mathcal{N}^d(0,I)$. As demonstrated in~\cite{neyshabur2017stabilizing}, a lower-dimensional DP-GAN is easier to train with far better convergent properties. The resulting generator $G_{ds}$ maps a $l$-dimensional ($l<d$) latent vector $z\in \mathbb{R}^l$ to a $d$-dimensional latent vector. For public data release, we randomly sample $z \sim \mathcal{N}^l(0,I)$ and synthesize synthetic image $G_{p}(G_{ds}(z))$. This whole process inherits the DP property from the DP-GAN, and the subsequent step of $G_{p}$ is a post-processing step that does not alter the DP property of the input, according to Theorem \ref{thm:post}. As such, the DPMI framework is $(\epsilon,\delta)$-DP if the latent space DP-GAN is $(\epsilon,\delta)$-DP, which was demonstrated in~\cite{abadi2016deep}.





\section{Experiments}
\label{sec:experiments}

In this section, we perform an ablation study on the model inversion, and compare the performances between the proposed DPMI scheme and DP-GAN at different privacy levels. To measure the performance of a GAN, we train a classifier using the output synthetic data and measure its accuracy on a hold-out test set of real images, \scc{as well as the quality of the synthetic images using established image quality and diversity metrics. This combined assessment approach} directly measures the validity of DPMI as a privacy-preserving data release mechanism for public machine learning. 

\subsection{Dataset Partition}

\scc{To make our proposed scheme broadly applicable to any GANs, we do not rely on any conditional inputs to the training. As such,} the generated synthetic data do not have labels and we need to extract a labeling dataset from our training dataset to train a labeling classifier. \scc{For all our experiments}, we separate the dataset properly into public and private domains. As the public dataset should be sufficiently different from the private dataset, we partition the dataset vertically, setting half of the classes as private and the other half as public. 

we first randomly select one-third of the training set as the labeling dataset $D_l$, which is used to train the labeling classifier. Then we divide the rest of the training set equally into $D_p$ and $D_s$, with a random half of the classes goes to $D_p$ and the other half goes to $D_s$. $D_p$ is treated as the public training set to train the public GAN and $D_s$ as the private training set is used for the model inversion process. The hold-out testing set contain test images from private classes only so as to test how well a classifier can predict private class labels. 
\subsection{Datasets and \cdj{Evaluation Metrics}}
\label{ssec:expermentsetting}
Three datasets are used in our experiments:
\begin{enumerate}
    \item CIFAR10~\cite{krizhevsky2009learning} contains 50,000 training and 10,000 testing color natural images in 10 classes of size 32$\times$32$\times$3. \cdj{In splitting between the public and private training sets, we randomly chose automobile, bird, cat, deer, and dog for $D_p$, and frog, horse, ship, truck, and airplane for $D_s$.}
    \item SVHN~\cite{netzer2011reading} is a color door-sign digit image dataset, containing 73,257 training images and 26,032 testing images of size 32$\times$32$\times$3. \cdj{The public set $D_p$ contains digits 1, 5, 7, 8 and 9 and $D_s$ contains digits 0, 2, 3, 4 and 6.}
    \item \cdj{Autism Spectrum Disorder Face (ASDF) dataset is a collection of children's facial landmark features extracted from videos captured for ASD screening~\cite{ozonoff2010prospective}. The dataset has 2,675,540 299-dimensional feature vectors consisting of facial landmarks, head pose, eye gaze and facial action units. Our target classification task is smile detection and the dateset is partitioned into 2,160,093 training and 515,447 testing samples. For the public/private split, we randomly separate the training set and use half of it as $D_p$ and use another half of it as $D_s$.}
\end{enumerate}



As these datasets post varying degrees of difficulty in their classification, different classification networks are used. 
For CIFAR10 and SVHN, we use convolutional neural network (CNN) classifiers with four convolution and max-pooling layers. 
\cdj{For ASDF, we use a hybrid structured classifier, containing both CNN and fully-connected layers.}

\scc{For performance measurements, we consider the accuracy of the trained classifier on the hold-out datasets, as well as image quality and diversity of the synthetic images.} \cdj{For CIFAR10 and SVHN, we use accuracy as the metric for classification performance as both datasets are balanced. The classification target is to predict the output labels for synthetic images. The ASDF dataset is an imbalanced dataset so we use the macro averages of precision, recall and F1-score as the evaluation metrics.} \cdj{For image quality and diversity measurements, we use Inception Score (IS) and Frechet Inception Distance (FID) based on a classifier trained on ImageNet to measure both synthetic image quality and diversity~\cite{borji2021pros}. These evaluation metrics are used only on CIFAR10 and SVHN as ASDF is not an image dataset.}

\subsection{Ablation study on Model Inversion}
To show the effectiveness of model inversion, we first evaluate model inversion under an ablation study. 
In this study, we first trained two non-DP GANs with generators $G_s$ and $G_p$ on the private and public training datasets $D_s$ and $D_p$ respectively. Three image classifiers are constructed based on the original private dataset $D_s$, synthetic images generated by $G_s$, and synthetic images generated through the model inversion process on the generator $G_p$. The test accuracy of classifiers trained on \cdj{the image} datasets are shown in Table~\ref{tab:cap}. \cdj{The macro-averages of precision, recall and F1-score of classifiers trained on the ASDF dataset are shown in Table~\ref{tab:asd}.}
\begin{table}[h!]
\begin{center}
\caption{Test accuracy of classifiers trained on $D_s$, synthetic data generated by $G_s$, and $G_p$ with model inversion} \label{tab:cap}
\begin{tabular}{|c|c|c|c|}
  \hline
  Dataset & $D_s$ & $G_s$ & $G_p$ + Model Inversion
  \\
  \hline
  CIFAR10 & 0.87 & 0.75 & 0.76 \\
  SVHN & 0.92 & 0.84 & 0.92 \\
  \hline
\end{tabular}
\end{center}
\end{table}
\begin{table}[h!]
\begin{center}
\caption{\cdj{Test macro average results of classifiers trained on $D_s$, synthetic data generated by $G_s$, and $G_p$ with model inversion for ASDF dataset}} \label{tab:asd}
\begin{tabular}{|c|c|c|c|}
  \hline
   & precision & recall & f1-score
  \\
  \hline
  $D_s$ & 0.66 & 0.69 & 0.67 \\
  $G_s$ &  0.63 & 0.67 & 0.63\\
  $G_p$ + Model Inversion & 0.65 & 0.68 & 0.66\\
  \hline
\end{tabular}
\end{center}
\end{table}

\scc{Tables \ref{tab:cap} and \ref{tab:asd} show that for both SVHN and ASDF, model inversion is able to achieve much better classification performance than synthetic images and rival those trained using the real private data. For CIFAR10, the gain is not as dramatic due to the relatively simplistic structure of our WGAN network.} To further examine how dissimilar the public and private datasets can be, we focus on the classification task on SVHN, and use CIFAR10 as the public $D_p$ and SVHN as the private $D_s$. In this experiment, the entire training and test datasets are used without any partitions. The same three configurations as the previous experiment are used and the results are shown in Table \ref{tab:transfer}. Model inversion method has a seven percentage point higher accuracy than using a private GAN and only a three percentage point lower than using the real SVHN training images. 
\begin{table}[h!]
\begin{center}
\caption{Test accuracy of classifiers trained on synthetic data with SVHN as $D_s$ and CIFAR10 as $D_p$} \label{tab:transfer}
\begin{tabular}{|c|c|}
  \hline
  Method & Accuracy\\
  \hline
  $D_s$ & 0.96 \\
  SVHN trained $G_s$ & 0.86 \\
  CIFAR10 trained $G_p$ + Model Inversion & 0.93 \\
  \hline
\end{tabular}
\end{center}
\end{table}

\subsection{Comparison with DP-GAN}
\scc{In this section, we compare the DPMI scheme with the DP-GAN implementation from \cite{xie2018differentially}. Specifically, we compare the accuracy of the classifiers trained using the synthetic data from DPMI and DP-GAN, and use the synthetic data generated by the private $G_s$ as the idealistic upper bound.} 

\begin{figure}[h!]
\centering
\centering
\includegraphics[width=7.0cm]{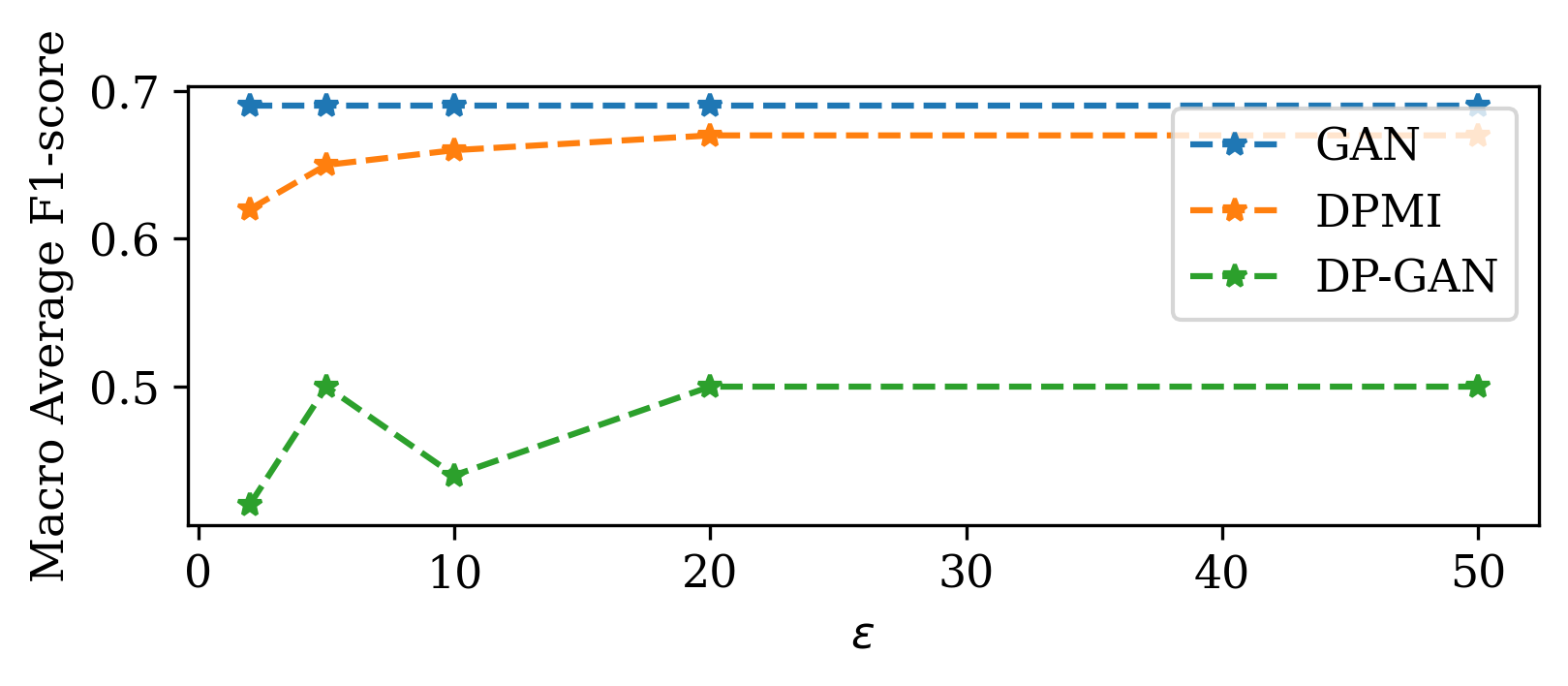}
\caption{\cdj{The F1-score of the classification for ASDF with different $\epsilon$.} }
\label{fig:OpenFace}
\end{figure}

\begin{figure}[h!]
\centering
\includegraphics[width=7.0cm]{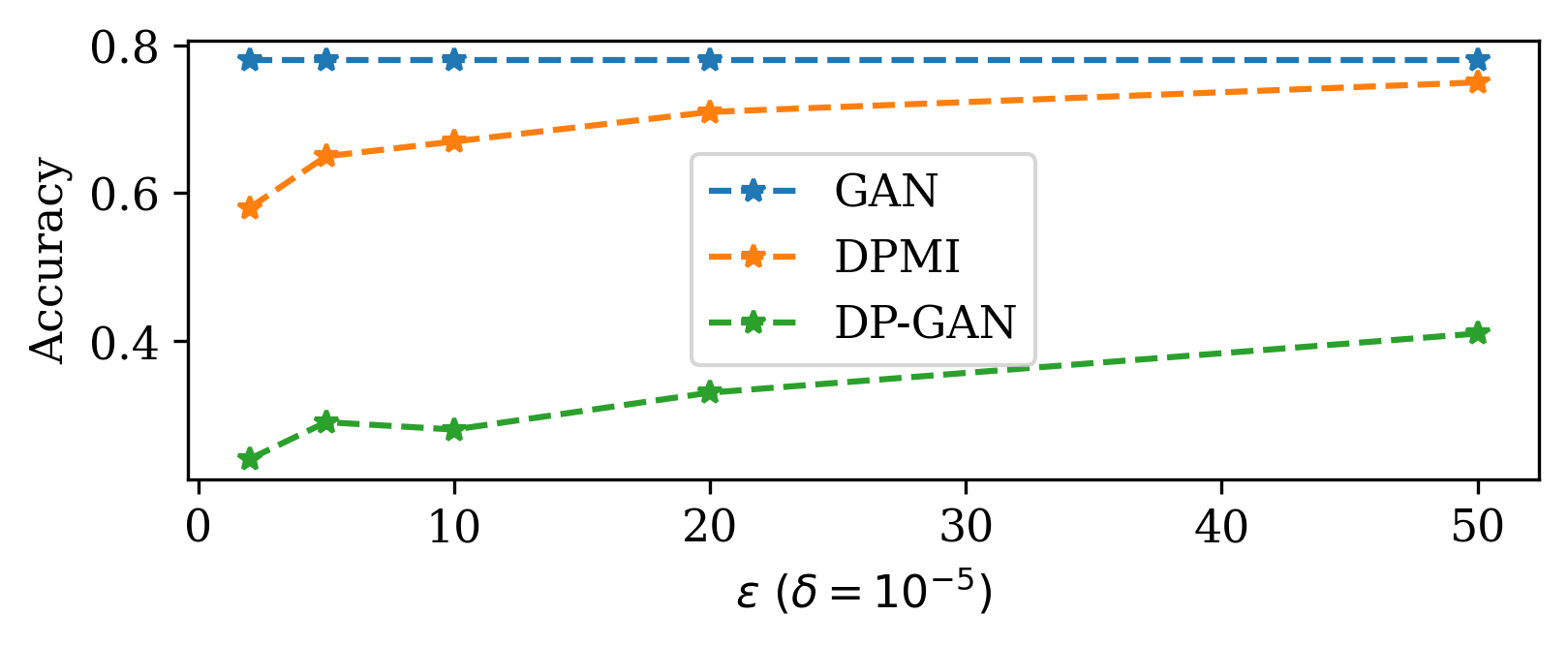}
\centerline{\scriptsize{(a) CIFAR10}}
\centering
\includegraphics[width=7.0cm]{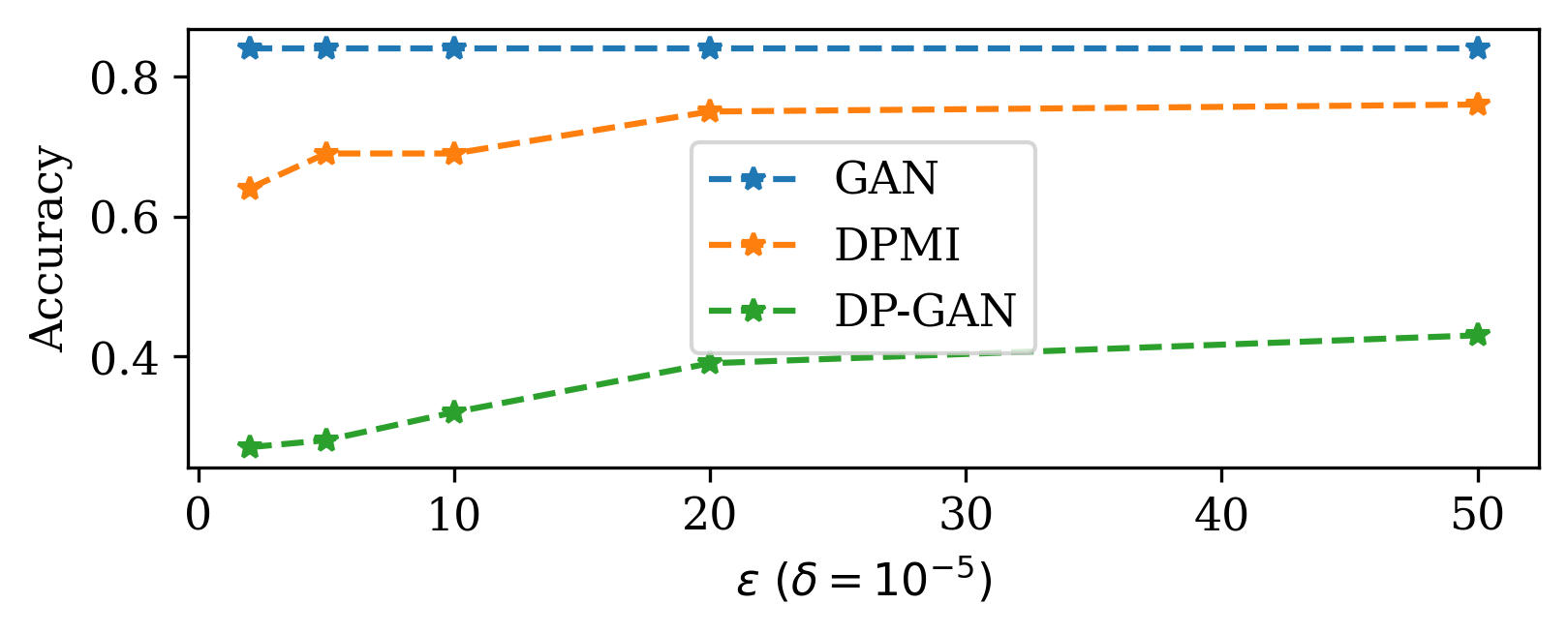}
\centerline{\scriptsize{(b) SVHN}}
\centering
\caption{The accuracy of the classification task for CIFAR10 and SVHN with different $\epsilon$. }
\label{fig:dpmi_results}
\end{figure}

Figures \ref{fig:OpenFace} and \ref{fig:dpmi_results} show the trade-off between testing accuracy and the privacy level $\epsilon$ for the three datasets. We find accuracy improvements from 34\% (0.75 vs 0.41, $\epsilon=50$) to 39\% (0.67 vs 0.28, $\epsilon=10$) on CIFAR10. For SVHN, our method has improvements from 33\% (0.76 vs 0.43, $\epsilon=50$) to 36\% (0.75 vs 0.39, $\epsilon=20$). \cdj{For ASDF, our method has improvements from 15\% (0.65 vs 0.50, $\epsilon=5$) to 22\% (0.66 vs 0.44, $\epsilon=10$). In fact, as shown Figure \ref{fig:OpenFace}, DPMI achieves similar performance as the ideal situation while DP-GAN simply does not converge as the results are mostly random guess on the smile/non-smile detection.} 

\begin{figure*}[h!]
\begin{multicols}{3}
    \includegraphics[width=\linewidth]{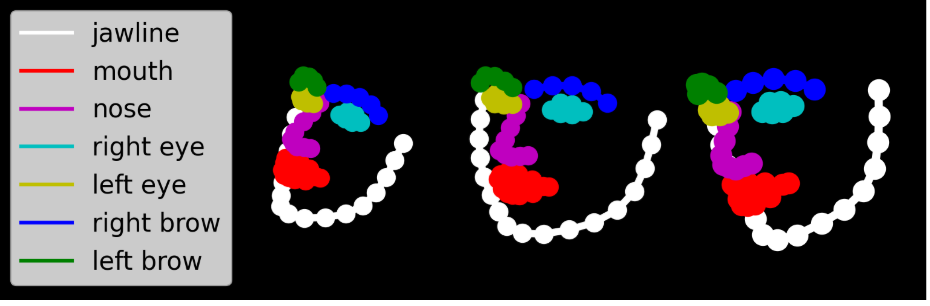}\par
    \centerline{\scriptsize{(a) real}}
    \includegraphics[width=\linewidth]{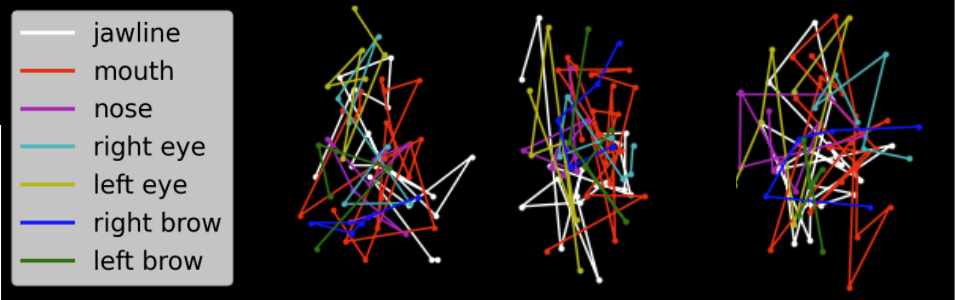}\par
    \centerline{\scriptsize{(b) DP-GAN ($\epsilon=10$)}}
    \includegraphics[width=\linewidth]{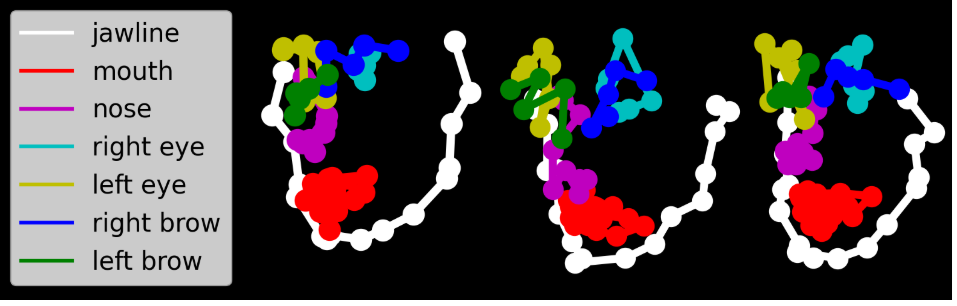}\par
    \centerline{\scriptsize{(b) DPMI ($\epsilon=10$)}}
\end{multicols}
\vspace*{-7mm}
\caption{Visual Results on ASDF dataset.}
\label{fig:ASDF}
\end{figure*}

\begin{figure}[h!]
\centering
\includegraphics[width=7.0cm]{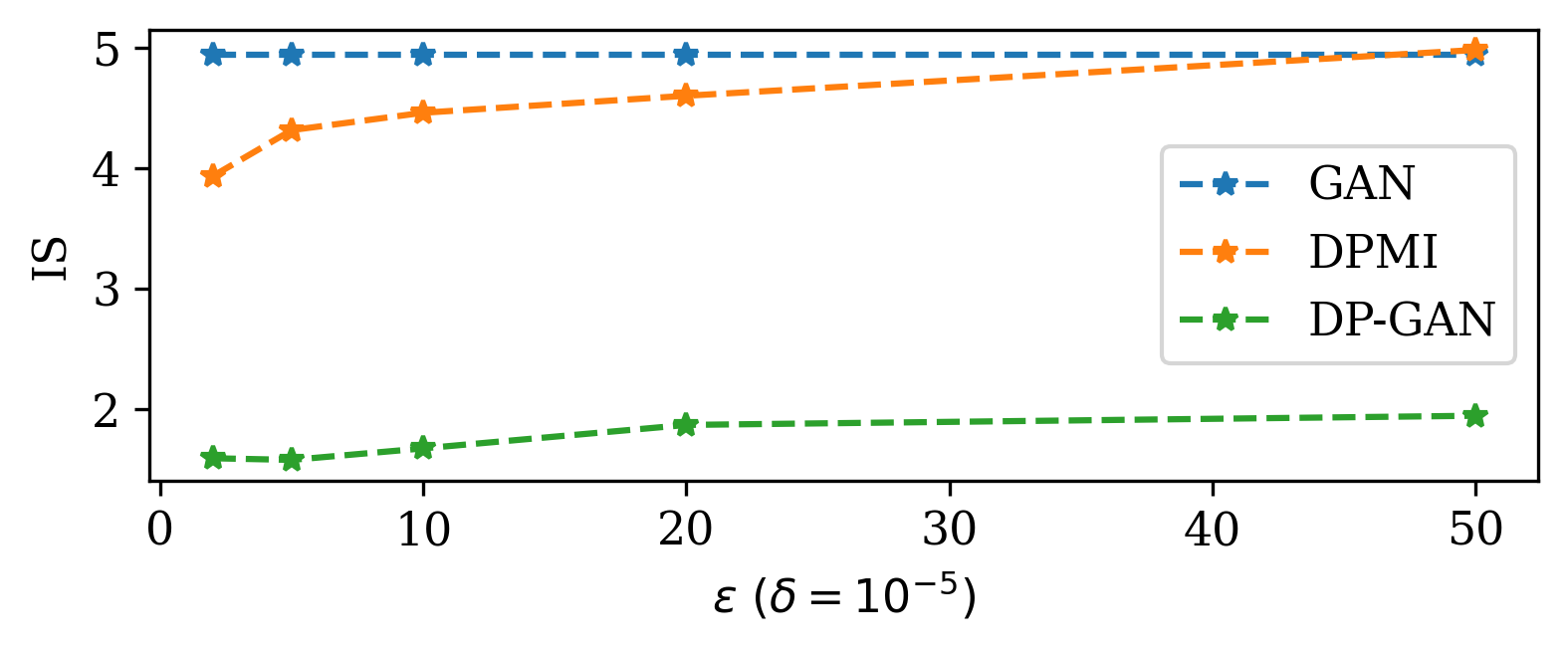}
\centerline{\scriptsize{(a) CIFAR10}}
\centering
\includegraphics[width=7.0cm]{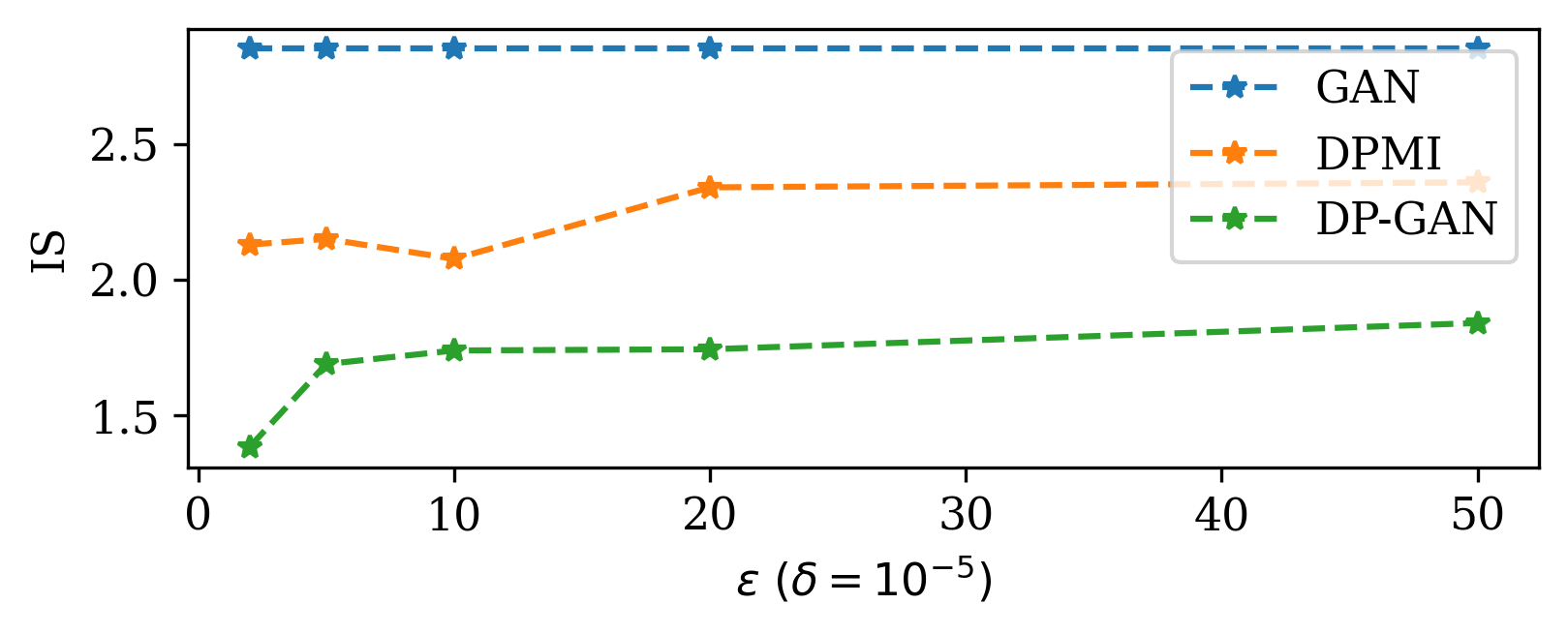}
\centerline{\scriptsize{(b) SVHN}}
\caption{\cdj{The Inception Score across different datasets with different $\epsilon$.} }
\label{fig:IS}
\end{figure}

\begin{figure}[h!]
\centering
\includegraphics[width=7.0cm]{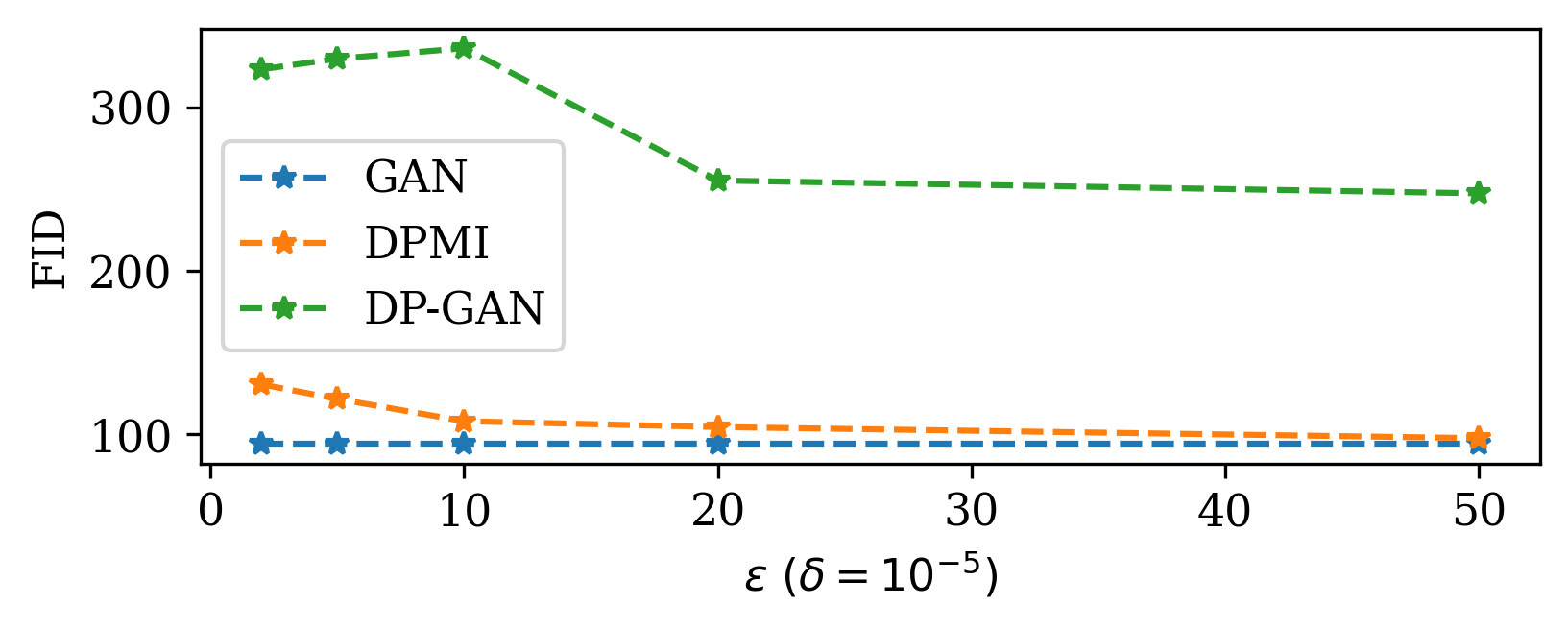}
\centerline{\scriptsize{(a) CIFAR10}}
\centering
\includegraphics[width=7.0cm]{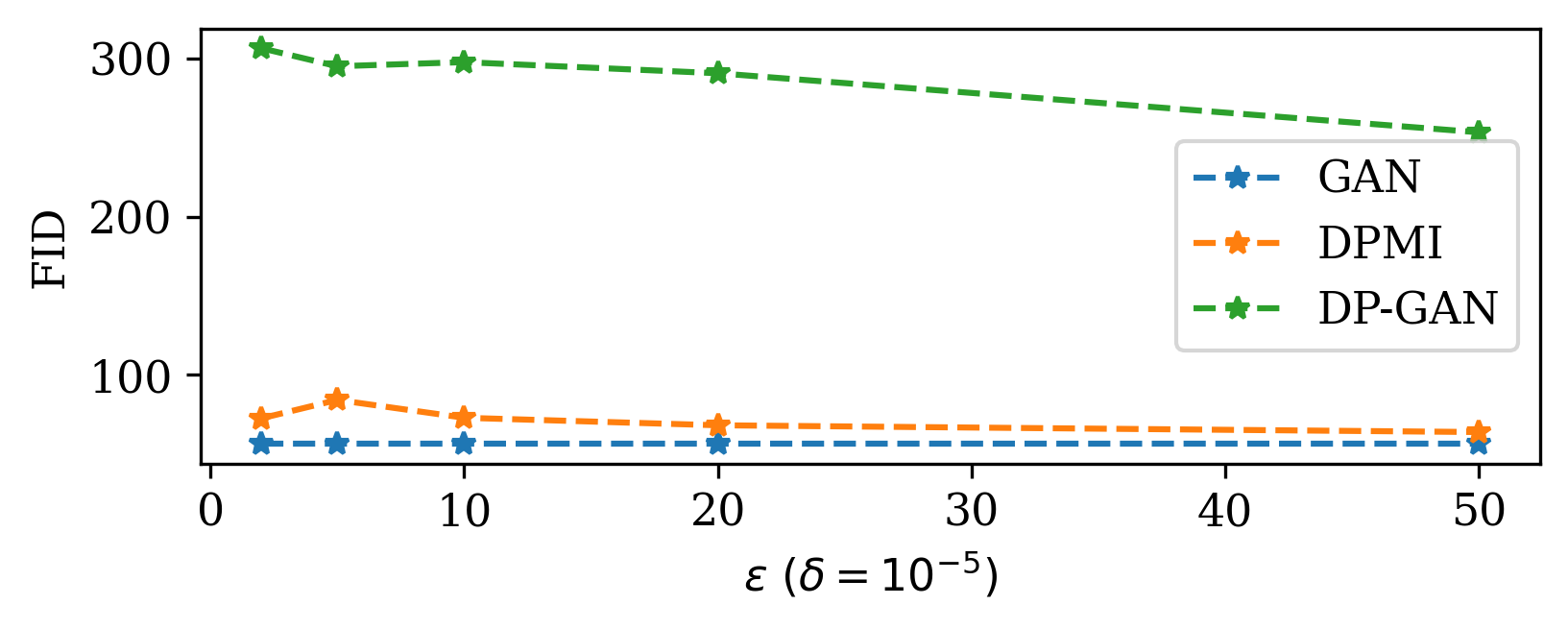}
\centerline{\scriptsize{(b) SVHN}}
\caption{\cdj{The Frechet Inception Distance across different datasets with different $\epsilon$.} }
\label{fig:FID}
\end{figure}

\cdj{Next, we consider the image quality and diversity among all the synthetic images. In Figure \ref{fig:IS} and \ref{fig:FID}, we compare the DPMI's performance to DP-GAN using Inception Score (IS) and Frechet Inception Distance (FID). A larger IS indicates the synthetic images have better quality and diversity. Figure \ref{fig:IS} shows that DPMI has higher IS than DP-GAN when they guarantee the same amount of privacy for both CIFAR10 and SVHN. FID is based on the feature distance between synthetic images and real testing images with the same labels. A smaller FID value means the synthetic images are closer to the real ones. As shown in Figure \ref{fig:FID}, DPMI has smaller FID values than DP-GAN under the same $\epsilon$ for CIFAR10 and SVHN. Note that since our private datasets only contains images from half of all the classes, the IS and FID values are generally worse than the state-of-the-art results. 
However, the relative comparisons using IS and FID sufficiently demonstrate that DPMI has better performance than the DP-GAN under the same privacy guarantee. }

\scc{As ASDF is not an image dataset, we are unable to compute neither IS nor FD. Figure \ref{fig:ASDF} shows a few visual examples from the original dataset and the synthetic ones from both DP-GAN and DPMI using $\epsilon=10$. Note that only the facial landmarks are displayed while the rest of the feature vectors do not have any reasonable methods for visualization. Nevertheless, it is quite clear that DPMI is able to produce reasonable landmarks while DP-GAN produces random results due to poor convergence of the training.} 

\section{Conclusions}
\label{sec:conclusion}
\scc{In this paper, we have introduced DPMI, a new differentially private generative framework of releasing synthetic private data by applying model inversion to map the real private data to the latent space via a public generator, followed by a lower-dimensional DP-GAN. Using two standard image datasets and one privacy-sensitive medical dataset, we have shown that model inversion process can improve the quality of the synthetic images and the lower-dimensional DP-GANs are able to produce better performance than the state-of-the-art DP-GAN in classification performance, synthetic image quality and diversity. Further investigation will focus on how different structures of GANs would affect the DPMI process and the end-to-end training of the model inversion process and the low dimensional DP-GAN.} 

\section{Acknowledgements}
\label{sec:acknowledgements}
Research reported in this publication was supported by the National
Institutes of Health, United States of America under award number
R01MH121344-01 and the Child Family Endowed Professorship.

\bibliographystyle{IEEEtran}
\bibliography{wifs2021}
\end{document}